\documentclass[onecolumn,conference]{IEEEtran}
\usepackage[T1]{fontenc}
\usepackage[latin9]{inputenc}
\usepackage{array}
\usepackage{booktabs}
\usepackage{multirow}
\usepackage{graphicx}
\usepackage{xcolor}
\usepackage{hyperref}
\usepackage{rotating}
\usepackage{tikz}
\usepackage{graphics}
\usepackage{enumitem}
\usepackage{balance}
\usepackage{amsmath}
\usepackage{comment}

\usepackage{xcolor,colortbl}
\newcommand{\tops}[1]{\cellcolor{gray}\textcolor{white}{{{#1}}}}
\newcommand{\baseline}[1]{\cellcolor{black}\textcolor{white}{{{#1}}}}
\hypersetup{pdftitle={Your Title},
 pdfauthor={Your Name},
 pdfpagelayout=OneColumn, pdfnewwindow=true, pdfstartview=XYZ, plainpages=false}

\makeatletter

\usepackage[caption=false,font=footnotesize]{subfig}

\makeatother

\begin{document}

\title{How to Identify Good Superpixels for Deforestation   Detection on Tropical Rainforests}

\author{
\IEEEauthorblockN{Isabela Borlido${}^{+}$, Eduardo Bouhid${}^{*}$,  Victor Sundermann${}^{+}$, Hugo Resende${}^{*}$\\ Alvaro Luiz Fazenda${}^{*}$,  Fabio Faria${}^{*}$,  Silvio Jamil F. Guimar\~aes${}^{+}$}
\\

\IEEEauthorblockA{${}^{*}$
\textit{Institute of Science and Technology} --
\textit{Universidade Federal de S\~{a}o Paulo} -- 
 S\~{a}o Jos\'{e} dos Campos, Brazil\\
 \{ebneto, hresende, alvaro.fazenda, ffaria\}@unifesp.br}

\IEEEauthorblockA{${}^{+}$
\textit{Computer Science Department} --
\textit{Pontif\'icia Universidade Cat\'olica de Minas Gerais} -- 
 Belo Horizonte, Brazil\\
 \{isabela.borlido, vgmsundermann\}@sga.pucminas.br,  sjamil@pucminas.br }
}

\maketitle

\begin{abstract}
\textcolor{black}{The conservation of tropical forests is a topic of significant social and ecological relevance due to their crucial role in the global ecosystem. Unfortunately, deforestation and degradation impact millions of hectares annually, requiring government or private initiatives for effective forest monitoring. However, identifying deforested regions in satellite images is challenging due to data imbalance, image resolution, low-contrast regions, and occlusion. Superpixel segmentation can overcome these drawbacks, reducing workload and preserving important image boundaries. However, most works for remote sensing images do not exploit recent superpixel methods. In this work, we evaluate 16 superpixel methods in satellite images to support a deforestation detection system in tropical forests. We also assess the performance of superpixel methods for the target task, establishing a relationship with segmentation methodological evaluation. According to our results, ERS, GMMSP, and DISF perform best on UE, BR, and SIRS, respectively, whereas ERS has the best trade-off with CO and Reg. In classification, SH, DISF, and ISF perform best on RGB, UMDA, and PCA compositions, respectively. According to our experiments, superpixel methods with better trade-offs between delineation, homogeneity, compactness, and regularity are more suitable for identifying good superpixels for deforestation detection tasks.}

\end{abstract}

\section{Introduction} 
Tropical rainforests play an essential role in preserving the global ecosystem, being vital for maintaining life and environmental balance on the planet. Endowed with rich biodiversity, these ecosystems harbor an immense variety of plant and animal species, many of which are exclusive to these regions. Additionally, tropical forests have a significant impact on climate regulation, absorbing substantial amounts of carbon from the atmosphere. The Brazilian Legal Amazon, in particular, represents a natural heritage of immeasurable value; however, it faces increasing threats from human activity, such as deforestation and degradation~\cite{lapola2023drivers, reis2023econometric}. In this sense, it is necessary to propose methodologies for the preservation of these forests, both to protect local biodiversity and ecosystems, and to mitigate the effects of global-scale climate change.

\begin{figure}[ht!]
    \centering
    \setlength{\tabcolsep}{0pt}
    \def\arraystretch{0}
    \begin{tabular}{cc}
         \includegraphics[width=0.5\linewidth]{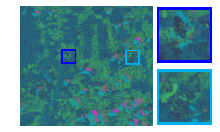} & \includegraphics[width=0.5\linewidth]{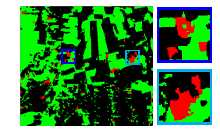} \\ 
         (a) Original Image & (b) Ground-truth \\
         \includegraphics[width=0.5\linewidth]{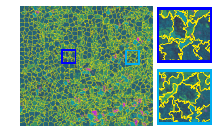} & \includegraphics[width=0.5\linewidth]{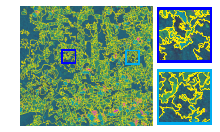} \\ 
         (c) SLIC segmentation & (d) DISF segmentation \\
    \end{tabular}
    \caption{Example of satellite image for deforestation. In (a) the original image and (b) the ground truth mask, in which forest regions are in green, recent deforestation in red, and old deforestation in black. Most classification methods incorporate a (c) SLIC segmentation to reduce workload, while recent methodologies, such as (d) DISF, remain unexplored.}
    \label{fig:intro:spx}
\end{figure}

In the literature, a variety of efforts to monitor and measure deforestation in the Amazon have been proposed, ranging from government programs to research projects that use machine learning \textcolor{black}{models}, image processing techniques, and citizen science methodologies. Government monitoring programs, such as the Program for Monitoring Deforestation in the Legal Amazon by Satellite (PRODES)~\cite{inpe-prodes}, provide essential data such as the annual deforestation rate and georeferenced maps, highlighting deforested areas. 
On the other hand, projects like ForestEyes~\cite{DALLAQUA2021422,dallaquaGRSL,10.1145/3653319}, which involve the participation of non-specialized volunteers on a citizen science platform to classify image segments \textcolor{black}{(so-called here by superpixels)} from Landsat-8 satellite covering forest and deforested areas. The quality of these image segments is crucial to maximizing \textcolor{black}{classification accuracy}, especially in the case of deforested areas. After this manual classification stage, these segments are used as training data for machine learning models, aiming at the automated identification of deforestation hotspots in the Amazon forest. 

\textcolor{black}{Image segmentation is an important resource for monitoring deforestation in remote sensing images, which is essential to using effective algorithms. Instead of using classical image segmentation methods, which could produce a few large regions, over-segmentation approaches (\textit{i.e.}, superpixel segmentation) have a better outcome, providing a compact image representation with low information redundancy while preserving object borders.} 
Superpixels are used in many applications, mainly as a pre-processing step. For instance, in land use classification of remote sensing images, modified versions of SLIC~\cite{achanta2012slic} method provide an initial segmentation based \textcolor{black}{on} remote sensing images~\cite{barbato2022unsupervised, li2022object}. In~\cite{li2022object}, SLIC is enriched with texture information, while in~\cite{barbato2022unsupervised} the SLIC's modification can handle hyperspectral information. 
Instead of performing an initial segmentation, in~\cite{zeng2023superpixel} a superpixel module is embedded into a CNN classifier that performs weakly supervised semantic segmentation to preserve boundary information. 
In contrast, a superpixel segmentation is used as post-processing in~\cite{yaobin2023Postprocessing} to reduce label fragmentation.


Despite \textcolor{black}{the vast} literature for superpixel segmentation, most methods are designed for RGB images, thereby not handling hyperspectral information~\cite{subudhi2021survey}. Also, remote sensing images impose different challenges from natural ones, such as low contrast regions and imbalanced data (Figure~\ref{fig:intro:spx}(a) and (b)). Additionally, most works for hyperspectral image analysis with superpixels use classical superpixel methods instead of recent approaches~\cite{subudhi2021survey} (Figure~\ref{fig:intro:spx}(c) and (d)). Recently, a survey for superpixel segmentation extensively evaluated several superpixel methods, most from recent literature, proving their benefits and drawbacks~\cite{barcelos2024review}. However, as far as we know, no such survey has been conducted to evaluate superpixel algorithms in remote-sensing applications.

In this regard, this work aims to fill this existing gap by providing a comprehensive analysis of 16 superpixel methods considering object delineation, color homogeneity, compactness, and regularity in remote sensing images for deforestation detection in tropical forests. We also thoroughly analyze superpixel methods as pre-processing for deforestation classification models. Furthermore, we correlate superpixel and classification assessments by ranking superpixel methods and identifying which ones are more suitable for the target task.
The paper is organized as follows. In Section~\ref{sec:superpixel}, discussions regarding the superpixel segmentation are given. Section~\ref{sec:experimental-setup} presents the experimental setup.
Section~\ref{sec:result} describes a quite extensive evaluation in terms of superpixel measures and deforestation classification. And finally, some conclusions and further works are drawn in Section~\ref{sec:conclusion}.

\section{Superpixel segmentation} 
\label{sec:superpixel}

An evaluation covering several recent and classical superpixel methods in natural and urban scenes was performed recently~\cite{barcelos2024review}. However, as depicted before, remote sensing images impose different challenges from those. For instance, the image quality is highly affected by weather and spatial resolution, and the image targets can be too small and low-contrast. Additionally, no superpixel segmentation is specifically designed for remote sensing images~\cite{subudhi2021survey}. We fill this gap by extensively evaluating 16 superpixel methods in remote sensing images for deforestation detection tasks. In the following, we briefly discuss the clustering \textcolor{black}{categories} of the superpixel methods~\cite{barcelos2024review} in this work. 
In neighborhood-based clustering methods, such as SLIC~\cite{achanta2012slic}, SCALP~\cite{GIRAUD-2018-SCALP}, and LSC~\cite{CHEN-2017-LSC}, superpixels conquer pixels based on a similarity measure restricted to a maximum spatial distance from some reference image point and a merging step is required to ensure connectivity. Conversely, dynamic-center-update clustering methods, such as SNIC~\cite{ACHANTA-2017-SNIC} and DRW~\cite{KANG-2020-DRW}, dynamically update the features of superpixel centers \textcolor{black}{through iterations, relying on them to conquer pixels.}

Boundary evolution-based clustering methods, such as 
ETPS~\cite{YAO-2015-ETPS}, 
perform an initial segmentation (usually in a grid) and iteratively update pixel blocks at the superpixel borders in a coarse-to-fine strategy. Conversely, path-based methods, such as ERGC~\cite{BUYSSENS-2014-ERGC}, ISF~\cite{VARGAS-2019-ISF}, RSS~\cite{CHAI-2020-RSS}, DISF~\cite{BELEM-2020-DISF}, 
and SICLE~\cite{BELEM-2022-SICLE}, perform a sampling strategy to initialize tree roots that iteratively conquer pixels with a path-based cost \textcolor{black}{function.} 
\textcolor{black}{Some superpixel methods employ deep neural networks to perform soft or hard pixel-superpixel assignment~\cite{barcelos2024review}, }
such as SSFCN~\cite{YANG-2020-SSFCN}. 
Hierarchical clustering methods, such as SH~\cite{WEI-2018-SH}, create a hierarchy from which one may compute many superpixel \textcolor{black}{scales.} 
The data distribution-based clustering methods, like GMMSP~\cite{BAN-2018-GMMSP}, assume that features in image pixels follow a specific distribution (\textit{e.g.}, the Gaussian one). Finally, graph-based clustering methods, such as ERS~\cite{LIU-2011-ERS}, perform superpixel segmentation based on graph topology. In addition to these methods, we also evaluate a GRID segmentation as a baseline. Other clustering categories for superpixel methods are covered in~\cite{barcelos2024review}.

\section{Experimental setup} \label{sec:experimental-setup}

In this study, the experimental dataset is composed of nine distinct Landsat-8 satellite multispectral images (7 channels) containing regions labeled as ``forest" and ``non-forest" provided by PRODES (ground-truth), operated by the United States Geological Survey, during July, August, September, and October 2022, covering the Xingu River Basin\footnote{Xingu River Basin - \url{https://maps.app.goo.gl/n9cR7s5ZNVtYrSqS7}} region, totaling over 8,514 $km^2$ of the Brazilian Legal Amazon. To establish which superpixel methods are more suitable for the deforestation detection task, we use the superpixel benchmark in~\cite{barcelos2024review} to compare $16$ methods -- including a grid segmentation baseline -- in nine available images. Since most superpixel \textcolor{black}{methods 
handle} multispectral \textcolor{black}{images, 
employ} the framework outlined in~\cite{neto2024satellite} for band selection. \textcolor{black}{Therefore, 
we} adopted three baseline compositions: (1) RGB channels extracted from LandSat-8 bands (B4B3B2); (2) three principal components extracted from Principal Component Analysis (PCA); and (3) Best UMDA Individual composed of four LandSat-8 bands (B4B3B1B6) provided by an evolutionary algorithm (Univariate Marginal Distribution algorithm -- UMDA~\cite{muhlenbein1996recombination}).

\textcolor{black}{In assessing superpixel segmentation, we use the following metrics (see~\cite{barcelos2024review} for more information): \textit{Boundary Recall} (BR)~\cite{martin2004learning} and \textit{Undersegmentation Error} (UE)~\cite{neubert2012superpixel} for object delineation, \textit{Explained Variation} (EV)~\cite{moore2008superpixel} and \textit{Similarity between Image and Reconstruction from Superpixels} (SIRS)~\cite{barcelos2022improving} for color homogeneity, and \textit{Compactness Index} (CO)~\cite{schick2012measuring} for compactness. 
Due to the sensor's spatial resolution used for image acquisition by PRODES, our ground-truth can only detect deforested areas exceeding $70$ pixels in size. Maintaining superpixels of similar sizes holds significance, particularly for Citizen Science initiatives~\cite{DALLAQUA2021422}. 
Therefore, we also assess \textit{Regularity} (Reg) by measuring the standard deviation of superpixel areas to determine how much their sizes differ in segmentation. }
\textcolor{black}{To correlate} superpixel and classification assessments, our experiments are organized in two parts: (1) a comparative analysis of the superpixel methods, which evaluates segmentation results within $1,000$ and $10,000$ superpixels and post-process them to guarantee connectivity \textcolor{black}{and a} minimum size of $70$ pixels for each superpixel; and (2) a comparative analysis of these methods for deforestation detection tasks, 
adopting as baseline method the experimental protocol used in~\cite{neto2024satellite}. 
\textcolor{black}{For superpixel segmentation, we use the parameters recommended by the original authors of each method.}

\section{Results and Discussion}\label{sec:result}

\subsection{Superpixel segmentation}

\begin{figure}[t!]
    \centering
    \includegraphics[width=0.75\linewidth]{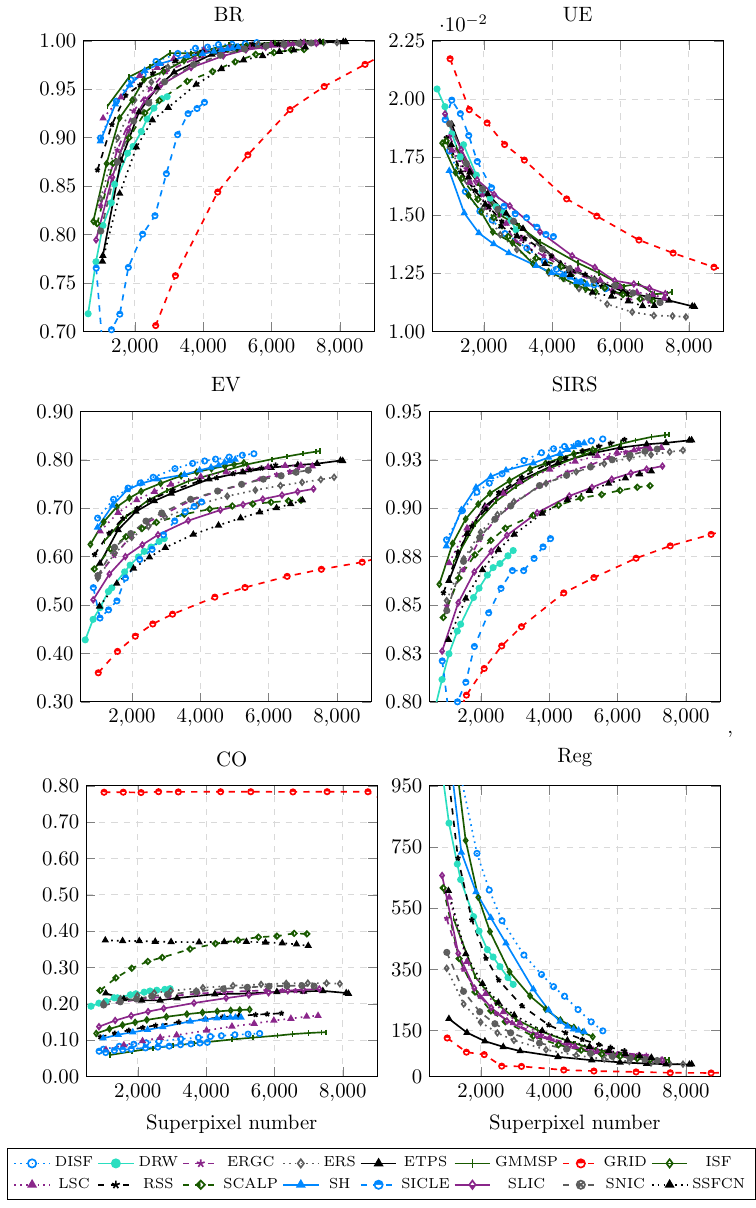}
        
    \caption{Superpixel results for object delineation (BR and UE) on top, color homogeneity (EV and SIRS) in the middle, and compactness (CO) and regularity (Reg) at the bottom.}
    \label{fig:res:delineation_homogeneity}
\end{figure}

In this section, we analyze the performance of superpixel \textcolor{black}{methods 
with} the deforested regions as ground-truth. 
Figure~\ref{fig:res:delineation_homogeneity} presents, from top to bottom, the results for delineation (BR and UE), color homogeneity (EV and SIRS), compactness (CO), and regularity (Reg). Higher is better in BR, EV, SIRS, and CO, while the opposite is true for UE and Reg. 
As it is possible to observe, GMMSP, LSC, DISF, SH, and RSS methods have the \textcolor{black}{best delineation (UE and BR)}, while the \textcolor{black}{methods GRID and SICLE} perform poorly. Additionally, \textcolor{black}{the deep-based method SSFCN struggles to delineate boundaries between recent deforestation and forest. 
Although ERS does not achieve the best BR, it has the lower UE}. Additionally, methods with better delineation also produce more homogeneous superpixels. 
As one may see in Figure~\ref{fig:res:delineation_homogeneity}, overall, methods with more compact superpixels have poorer delineation. Among these, SCALP has better delineation, \textcolor{black}{whereas SICLE performs} poorly in both evaluations. As expected, GRID produces the most regular superpixels, \textcolor{black}{followed by ETPS}. The regularity is overall low --- \textit{i.e.}, the variation among superpixel size is high, but it improves when the number of superpixels increases. \textcolor{black}{However, SH, ISF, RSS, and DRW} have worse regularity than other methods. 

\textcolor{black}{We compute a score for each superpixel method to provide a ranking strategy covering the important superpixel properties.} First, we compute the mean result across different numbers of superpixels for each method and measure. 
Then, we sort superpixel methods for each evaluation measure, generating rank values, such that the best evaluation receives rank 1 and equal evaluation results have equal ranks. 
Finally, the score of a superpixel method is given by its mean rank. Since EV is sensitive to textures and SIRS also evaluates color homogeneity, we remove EV from the \textcolor{black}{ranking}. 
The final rank and scores are shown in Table~\ref{tab:res:rank:spx}, in which ERS, ETPS, and GMMSP had the three best scores. ERS has the lowest leakage (UE) and balanced performance in the remaining assessment measures. ETPS has good Regularity (Reg) and color homogeneity (SIRS), while GMMSP has the best delineation (BR) but worse compactness (CO). 

\begin{table}[t!]
\centering
\caption{Rank for superpixel evaluation measures and the final score for each method (lower values mean better performance).}
\label{tab:res:rank:spx}
\resizebox{0.6\linewidth}{!}{
\begin{tabular}{@{}clcccccc@{}}
\toprule
\textbf{Rank} & \textbf{Method} & \textbf{BR} & \textbf{UE} & \textbf{SIRS} & \textbf{CO} & \textbf{Reg} & \textbf{Score} \\ \midrule
1 & ERS \cite{LIU-2011-ERS} & 6 & 1 & 8 & 4 & 3 & 4.4 \\
2 & ETPS \cite{YAO-2015-ETPS} & 8 & 6 & 4 & 6 & 2 & 5.2 \\
3 & GMMSP \cite{BAN-2018-GMMSP} & 1 & 4 & 2 & 15 & 5 & 5.4 \\
4 & LSC \cite{CHEN-2017-LSC} & 2 & 5 & 5 & 13 & \textbf{7} & 6.4 \\
5 & SH \cite{WEI-2018-SH} & 4 & 2 & 3 & 12 & 12 & 6.6 \\
6 & ERGC \cite{BUYSSENS-2014-ERGC} & 7 & 8 & 9 & 7 & 6 & 7.4 \\
7 & SNIC \cite{ACHANTA-2017-SNIC} & 9 & 10 & 10 & 5 & 4 & 7.6 \\
8 & SSFCN \cite{YANG-2020-SSFCN} & 13 & 3 & 13 & 2 & 10 & 8.2 \\
8 & SCALP \cite{GIRAUD-2018-SCALP} & 12 & 7 & 11 & 3 & 8 & 8.2 \\
8 & DISF \cite{BELEM-2020-DISF} & 3 & 9 & 1 & 14 & 14 & 8.2 \\
11 & RSS \cite{CHAI-2020-RSS} & 5 & 12 & 6 & 11 & 11 & 9.0 \\
12 & GRID & 16 & 14 & 14 & 1 & 1 & 9.2 \\
13 & ISF \cite{VARGAS-2019-ISF} & 10 & 11 & 7 & 10 & 13 & 10.2 \\
14 & SLIC \cite{neto2024satellite} & 11 & 13 & 12 & 9 & 9 & 10.8 \\
15 & DRW \cite{KANG-2020-DRW} & 14 & 16 & 15 & 8 & 15 & 13.6 \\
16 & SICLE \cite{BELEM-2022-SICLE} & 15 & 15 & 16 & 16 & 16 & 15.6 \\ \bottomrule
\end{tabular}
}
\end{table}

\begin{table}[ht!]
\centering
\caption{\textcolor{black}{Classification's balanced accuracy for RGB, PCA, and UMDA and the final score for each method (lower values mean better performance). The best results and baseline are in gray and black, respectively.}}
\resizebox{0.6\linewidth}{!}{
\begin{tabular}{clcccc}
\toprule
{\textbf{Rank}} & {\textbf{Method}} & \textbf{RGB} & \textbf{PCA} & \textbf{UMDA} & {\textbf{Score}}\\ \midrule 
1 & SH~\cite{WEI-2018-SH} & \tops{84.65} & 88.42 & 86.06 & 1.7\\ 
2 & DISF~\cite{BELEM-2020-DISF} & 84.56 & 87.85 & \tops{86.10} & 4.0\\ 
3 & ISF~\cite{VARGAS-2019-ISF} & 84.53 & \tops{88.88} & 85.16 & 4.7\\ 
4 & RSS~\cite{CHAI-2020-RSS} & 84.22 & 88.26 & 84.86 & 6.0\\ 
5 & SNIC~\cite{ACHANTA-2017-SNIC} & 83.44 & 87.87 & 85.62 & 6.7\\ 
6 & DRW~\cite{KANG-2020-DRW} & 84.07 & 87.78 & 85.60 & 7.3\\ 
7 & SICLE~\cite{BELEM-2022-SICLE} & 84.13 & 86.86 & 85.80 & 7.7\\ 
8 & LSC~\cite{CHEN-2017-LSC} & 83.07 & 87.94 & 85.40 & 8.0\\ 
9 & ERS~\cite{LIU-2011-ERS} & 82.86 & 88.02 & 85.33 & 8.3\\ 
10 & GMMSP~\cite{BAN-2018-GMMSP} & 83.86 & 87.79 & 85.16 & 8.7\\ 
\baseline{10} & \baseline{SLIC~\cite{neto2024satellite}} & \baseline{83.35} & \baseline{88.10} & \baseline{84.54} & \baseline{8.7}\\ 
12 & SSFCN~\cite{YANG-2020-SSFCN} & 83.49 & 87.43 & 85.24 & 9.7\\ 
13 & SCALP~\cite{GIRAUD-2018-SCALP} & 83.08 & 87.86 & 84.44 & 11.0\\ 
14 & ETPS~\cite{YAO-2015-ETPS} & 82.43 & 87.50 & 84.54 & 13.3\\ 
15 & ERGC~\cite{BUYSSENS-2014-ERGC} & 82.46 & 87.28 & 83.79 & 14.3\\ 
16 & GRID & 79.16 & 84.52 & 81.24 & 16.0\\ 
\bottomrule
\multicolumn{2}{r}{\textbf{Relative Gain \tiny{(Best  $\times$ SLIC)}}}& \textbf{1.57}  & \textbf{0.89} &   \textbf{1.84}   & - \\
\bottomrule
\end{tabular}
}
\label{tab:acc} 
\end{table}

\subsection{Classification}

The goal of these experiments is to verify whether the methodology used in 
superpixel segmentation also corresponds to the deforestation detection task, \textit{i.e.}, whether the best superpixel methods found previously correspond to the best in the target task. For this purpose, we adopted the same protocol used in~\cite{neto2024satellite}. Therefore, we split the dataset into 7 images for the training set and 2 for the test and set $6,000$ superpixels per image.
First, we extract features from superpixels using Haralick~\cite{HARALICK} descriptor, which codes textural properties into feature vectors for each image channel.  Next, we train and test a support vector machine (SVM~\cite{svm}) with superpixels from the train and test images, respectively. 
Subsequently, we train a Support Vector Machine (SVM)~\cite{svm} for each superpixel method with their superpixels from the training set. The classifiers are then assessed on the test set. To ensure fairness, the test set consists of SLIC superpixels extracted from the test images. 

Table~\ref{tab:acc} shows the classification results by the SVM model with Haralick descriptor for the $16$ superpixel methods. 
Furthermore, similar to~\cite{neto2024satellite}, three different image compositions (RGB, PCA, and UMDA) have been considered to represent the segments created by superpixel methods, and we adopt SLIC as the baseline. As one may see, SH, ISF, and DISF methods perform slightly better than the remaining methods for RGB, PCA, and UMDA image composition, respectively. When these methods are compared with the baseline method 
(SLIC), they achieved relative gains of $1.57\%$, $0.89\%$, and $1.84\%$. In addition, a final ranking list was created considering the same score procedure applied to superpixel evaluation in the previous experiment on the methods' performance in the three compositions. SH method achieved first place among all superpixel methods here compared.

\subsection{Discussion}\label{sec:discussion}

\begin{figure}[t!]
    \centering
    \includegraphics[width=1\linewidth, trim={1cm 0 0.2cm 0},clip]{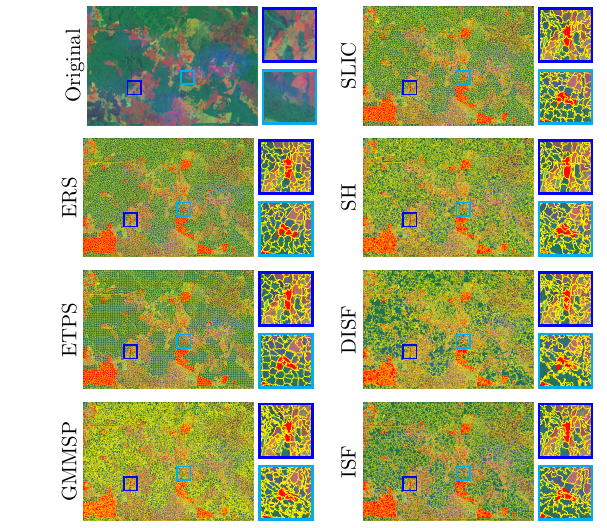}
    \caption{\textcolor{black}{Original image with PCA composition and segmentation with 6000 superpixels for SLIC, ERS, SH, ETPS, DISF, GMMSP, and ISF. In superpixel segmentations, the superpixel borders are indicated in yellow and recently deforested regions are colored in red.}}
    \label{fig:qualitative}
\end{figure}

\textcolor{black}{As one may see in Tables~\ref{tab:res:rank:spx} and~\ref{tab:acc}, the baseline SLIC is not optimal for either segmentation or classification task}. \textcolor{black}{Figure~\ref{fig:qualitative} presents a comparison of segmentation results of the classification baseline SLIC (top right) with the best three methods for segmentation (according to Table~\ref{tab:res:rank:spx}) and the best three for classification (according to Table~\ref{tab:acc}). As one may see, methods with superior object delineation (SH, DISF, and ISF) usually lead to better classification results, although producing less compact and less regular superpixels}. The top 4 methods in Table~\ref{tab:acc} produce highly homogeneous superpixels, with the first two offering better object delineation. 
\textcolor{black}{In terms of segmentation results, ERS has the best balance across evaluation measures and all methods.} 
Nevertheless, its classification performance \textcolor{black}{is worse than} DRW, which has poor segmentation performance. Similarly, GMMSP has excellent delineation, but its classification result is inferior to SICLE, which underperforms in all superpixel evaluation measures. 

\textcolor{black}{Considering the segmentation and classification results (Tables~\ref{tab:res:rank:spx} and~\ref{tab:acc}), one may observe that more homogeneous superpixels with better delineation generally have improved classification. However, in Citizen Science projects, more regular and compact superpixels are more friendly to an untrained user. Therefore, our results indicate that methods with better trade-offs between delineation, homogeneity, compactness, and regularity (ERS, ETPS, and GMMSP in Figure~\ref{fig:qualitative}) are more suitable for identifying good superpixels for deforestation classification in Citizen Science projects. Nonetheless, to improve classification performance in such projects, it may be worth exploring strategies to generate more compact and regular superpixels from methods with high delineation to maximize classification.}

\section{Conclusions}\label{sec:conclusion}

In this work, we evaluate 16 superpixel methods in Landsat-8 satellite images in terms of superpixel segmentation and deforestation classification. Since no superpixel method outperforms in all evaluations, we rank the methods' performance, identifying which ones are most suitable for each task. Considering the best ten methods in Tables~\ref{tab:res:rank:spx} and~\ref{tab:acc}, SH, ERS, LSC, GMMSP, and SNIC excel in both tasks, while DISF, ISF, and RSS have excellent classification, delineation, and color homogeneity, but low compactness and regularity. Our study shows that improved delineation and color homogeneity lead to better classification. For further works, we intend to explore strategies for adequate superpixel methods with high delineation to the ForestEyes~\cite{DALLAQUA2021422} project. Also, we aim to explore other image band selections in deforestation classification with these methods.

\section*{Acknowledgement}

The authors thank the PUC Minas, the CAPES (Finance Code 001), the CNPq ($\#$407242/2021-0 and $\#$306573/2022-9) and the FAPEMIG ($\#$APQ-01079-23). This research is part of INCT Future Internet for Smart Cities, funded by CNPq ($\#$465446/2014-0), and FAPESP ($\#$2014/50937-1, $\#$2015/24485-9, $\#$2017/25908-6, $\#$2018/23908-1, $\#$2019/26702-8, $\#$2023/00811-0, and $\#$2023/00782-0). Acknowledgements to the National Laboratory for Scientific Computing (LNCC/MCTI, Brazil) for providing HPC resources of the SDumont\footnote{SDumont - \url{http://sdumont.lncc.br}} supercomputer.


\end{document}